%% file: conference_101719.tex
\documentclass[a4paper, 10pt, conference]{IEEEtran}
\IEEEoverridecommandlockouts
\usepackage[a4paper, top=0.38in, bottom=1in, left=0.62in, right=0.62in]{geometry}
\usepackage{times}

\usepackage{cite}
\usepackage{amsmath,amssymb,amsfonts}
\usepackage{algorithmic}
\usepackage{graphicx}
\usepackage{textcomp}
\usepackage{xcolor}
\def\BibTeX{{\rm B\kern-.05em{\sc i\kern-.025em b}\kern-.08em
    T\kern-.1667em\lower.7ex\hbox{E}\kern-.125emX}}
\renewcommand\IEEEkeywordsname{Keywords}

\begin{document}

\title{Enhancing Deepfake Detection using SE Block Attention with CNN }

\makeatletter
\newcommand{\newlineauthors}{%
  \end{@IEEEauthorhalign}\hfill\mbox{}\par
  \mbox{}\hfill\begin{@IEEEauthorhalign}
}
\makeatother

\author{\IEEEauthorblockN{Subhram Dasgupta}
\IEEEauthorblockA{\textit{Department of Computer Science} \\
\textit{North Carolina A\&T State University}\\
Greensboro, USA \\
sdasgupta@aggies.edu}
\and
\IEEEauthorblockN{Janelle Mason}
\IEEEauthorblockA{\textit{Department of Computer Science} \\
\textit{North Carolina A\&T State University}\\
Greensboro, USA \\
jcmason@aggies.ncat.edu}
\and
\IEEEauthorblockN{Xiaohong Yuan}
\IEEEauthorblockA{\textit{Department of Computer Science} \\
\textit{North Carolina A\&T State University}\\
Greensboro, USA \\
xhyuan@ncat.edu}
\newlineauthors
\IEEEauthorblockN{Olusola Odeyomi}
\IEEEauthorblockA{\textit{Department of Computer Science} \\
\textit{North Carolina A\&T State University}\\
Greensboro, USA \\
otodeyomi@ncat.edu}
\and
\IEEEauthorblockN{Kaushik Roy}
\IEEEauthorblockA{\textit{Department of Computer Science} \\
\textit{North Carolina A\&T State University}\\
Greensboro, USA \\
kroy@ncat.edu}
}

\maketitle

\begin{abstract}
In the digital age, Deepfake present a formidable challenge by using advanced artificial intelligence to create highly convincing manipulated content, undermining information authenticity and security. These sophisticated fabrications surpass traditional detection methods in complexity and realism. To address this issue, we aim to harness cutting-edge deep learning methodologies to engineer an innovative deepfake detection model. However, most of the models designed for deepfake detection are large, causing heavy storage and memory consumption. In this research, we propose a lightweight convolution neural network (CNN) with squeeze and excitation block attention (SE) for Deepfake detection. The SE block module is designed to perform dynamic channel-wise feature recalibration. The SE block allows the network to emphasize informative features and suppress less useful ones, which leads to a more efficient and effective learning module. This module is integrated with a simple sequential model to perform Deepfake detection.  The model is smaller in size and it achieves competing accuracy with the existing models for deepfake detection tasks. The model achieved an overall classification accuracy of 94.14\% and AUC-ROC score of 0.985 on the Style GAN dataset from the Diverse Fake Face Dataset. Our proposed approach presents a promising avenue for combating the Deepfake challenge with minimal computational resources, developing efficient and scalable solutions for digital content verification.

\end{abstract}


\section{Introduction}
A Deepfake refers to digital content created by artificial intelligence that appears genuine to human observers. The term `deepfake' merges `deep learning' with `fake', indicating its origin from artificial neural networks, a subset of machine learning techniques. Deepfake are most often associated with the creation and alteration of human images, making them convincingly realistic through the manipulation of visual elements \cite{mirsky2021creation}.
Deepfake have been used to tarnish reputations, manipulate perceptions, and spread misinformation. The creation of Deepfake portrays individuals in a wrong way, which can lead to public outrage and misinformation campaigns that can potentially sway public opinion and change the political scene. Deepfake have created a landscape where seeing is no longer believing, instigating a critical public discourse on trust and authenticity in the digital age. The first time a Deepfake attack existed online was in 2017. It was used by a Reddit user named `Deepfake' used deep learning-based Deepfake methods to create pornographic content with swapped faces of celebrities and published the content online \cite{yu2021survey}. Over the past few years, technology has progressed and there are several examples where Deepfake have been used to commit fraudulent financial acts, promote fake news, and more. Therefore, the detection of manipulated content has become very important to help judge the authenticity of content. 
    
Deepfakes are generated using Generative Adversarial Networks (GANs) \cite{seow2022comprehensive}. A wide range of face manipulation approaches involve state-of-the-art methods in deep learning and computer vision. Generally, face forgery methods are largely classified into four types: Identity Swap, Face Reenactment, Attribute Manipulation and Entire Face Synthesis \cite{kong2022digital}. Identity swap, also called Face Swap is a method where the face of a source is swapped with a target \cite{huang2023implicit}. In the second type, Face Reenactment, the expression of the target face is manipulated instead of swapping the entire face \cite{mirsky2021creation}. Neural Textures is an example of this method. The third method, Attribute Manipulation, is used in the Faceapp application. Manipulation of attributes of a face (such as facial hair, length of hair, and other accessories are added like spectacles, sunglasses, cosmetic makeup, and more) is done in attribute manipulation \cite{naitali2023deepfake}. The latest and the most pristine form is the entire face synthesis method which uses different GAN architectures to construct manipulated faces. To tackle these Deepfake, many datasets have been made publicly available to develop deepfake detection methods. Faceforensics++ \cite{rossler2019faceforensics++} and Diverse Fake Face Dataset (DFFD) \cite{dang2020detection} are some example datasets. Zhu et al\cite{zhu2021attention}. designed a CNN with attention mechanism and Wodajo et al\cite{wodajo2021deepfake}. developed a convolutional vision transformer for the detection purpose. There are other methods that explore the area of implementing attention mechanisms for deepfake detection.
    
The Squeeze and Excitation (SE) block, proposed in \cite{hu2018squeeze}, introduces a building block to the CNNs. The SE block enhances image classification by focusing on informative features and suppressing irrelevant ones, improving discriminative capabilities. Its lightweight nature allows efficient integration into existing architectures, making it a versatile choice for various tasks. The adaptive feature recalibration helps increase classification accuracy. They were used at the ImageNet competition and helped to improve the results from the previous year by a significant amount. Furthermore, Deepfake image classification benefits greatly from the SE block`s ability to improve discriminative skills and concentrate on informative features because Deepfake images differ slightly from real ones, necessitating careful feature extraction for precise detection and classification.
    
The objective is to enhance the detection accuracy of the manipulated content. The integration of SE block into CNN architectures enhances the feature extraction capabilities of the model by dynamically recalibrating channel-wise feature responses. SE block enhances the model`s ability to distinguish between more and less informative features, leading to a more effective and efficient extraction of relevant features from the input data. In our research, we introduce a streamlined CNN architecture enhanced by a SE block. This novel integration of the SE block attention mechanism within the CNN framework, which demonstrates promising outcomes. Experimental findings reveal that our proposed model delivers comparable performance in detecting Deepfake relative to existing cutting-edge methodologies.
    
The contributions of this research are summarized as follows.
    \begin{itemize}
        \item We introduce a compact SE block-enhanced CNN specifically designed for Deepfake detection. 
        \item The SE block is employed to perform channel-wise feature extraction, improving the model’s efficiency and accuracy with minimal computational increase.
    \end{itemize}

In the next section of this paper, a literature review is provided. The remainder of the paper is as follows. Section 3 is the research methodology that explains the steps that were followed to conduct our research. The experiments that were conducted, as well as the results and analysis are discussed in Section 4. Section 5 concludes the research work and defines the future work.
\section{Related Works}
Deepfake represent a sophisticated digital phenomenon where artificial intelligence and machine learning technologies, particularly deep learning algorithms, are utilized to create or manipulate video and audio content with a high degree of realism. These convincing forgeries can have significant implications, ranging from misinformation and privacy breaches to security concerns, as they become increasingly indistinguishable from authentic content.
The following sections of this paper will examine different ways that have been created to detect Deepfake, highlighting the innovative approaches and methodologies employed to counteract the challenges posed by these hyper-realistic fabrications. 

In Section \ref{dfd}, a summary of existing deepfake detection methods is discussed, In Section \ref{seb}, SE block attention and its applications in deepfake detection are introduced.
\subsection{Deepfake Detection}\label{dfd}
Deep learning-based frameworks are widely used for deepfake detection. CNNs are the most commonly used model. Some deep learning-based models used for deepfake detection are XceptionNet, GoogleNet, VGG, ResNet, EfficiemtNet, MobileNet and more. Some Recurrent Neural Network (RNN) models are also used \cite{rana2022deepfake}. As the quality of Deepfake became better, new architectures with attention modules were introduced like Vision Transformers (ViT) \cite{dosovitskiy2020image}.
The detection of Deepfake is largely dependent upon the diminutive details of face images. The introduction of an attention mechanism for fake face detection has reaped benefits. Researchers have developed advanced detection approaches by combining the CNN and ViT architectures. EfficientNet B0 is integrated with ViT to create a model for deepfake detection \cite{coccomini2022combining}.
ViTs generally exhibit moderate performance on datasets of moderate size compared to CNNs. They need large-scale datasets to perform well. Convolutional Vision Transformers address the issue of getting the best out of both convolutional neural networks and Vision Transformers \cite{wu2021cvt}.  
\subsection{SE Block CNNs}\label{seb}
The concept of SE blocks in CNNs was first introduced in a submission for the ImageNet Large Scale Visual Recognition Challenge (ILSRVC) 2017 classification challenge \cite{hu2018squeeze}. In a traditional CNN like VGG16, the convolution process amalgamates both spatial and channel-wise information in local receptive fields to extract features. This allows the network to learn hierarchical feature representations crucial for various tasks like image classification and object detection \cite{simonyan2014very}. The SE block relies on its ability to adaptively recalibrate the channel-wise feature responses. It achieves this by specifically focusing on the interdependencies among the channels. By integrating this architecture, substantial enhancements in performance have been observed in various leading deep learning architectures, all while incurring only a minimal increase in computational demand \cite{hu2018squeeze}.

Roy et al \cite{roy20223d}. explored the usage of SE block architecture on the FaceForensics++ datasets for the detection of Deepfake, where it achieved better generalization compared to the CNN models used to conduct the experiments \cite{rossler2019faceforensics++}. 

In the following section, we present and explain our proposed architecture which utilizes the ability of the SE block attention mechanism on a simple sequential model. Satisfying results are shown in comparison to models that are data-hungry and computation-heavy for the state-of-the-art Deepfake dataset, (DFFD) \cite{dang2020detection}. A summary of the dee


\section{Proposed Architecture} \label{sec:proposed}
In our proposed architecture, we use a Sequential CNN with a SE block attention module \cite{hu2018squeeze}. These two components work together to improve the model`s performance. The following subsections provide a detailed description of each module and explain how the two modules are integrated to form our suggested solution.
\subsection{Simple CNN Module} \label{sec:cnn}
The CNN modules used in the proposed architecture are used to classify the image data or data that can be interpreted in the form of images. Keras, in the TensorFlow platform, is used to build these models. The model is sequential, essentially because the output of the previous layer is used as the input to the next layer. In the CNN modules, filters, also known as kernels, play a pivotal role in feature extraction from input images. These filters are small, but trainable matrices that convolve across the input image to perform element-wise multiplication with the part of the image they cover, summing up the results into a single output pixel in the feature map. This process allows the network to capture spatial hierarchies and patterns such as edges, textures, or more complex shapes in deeper layers. The \textit{3 X 3 filter} is used over \textit{1 X 1 filter} or \textit{5 X 5 filter} because of its efficiency in feature learning. The activation function, \textit{ReLU} or Rectified Linear Unit is used for its linear representation ability of non-linear data.
The \textit{ReLU} function is defined as:
\begin{equation}
    f(x) = \max(0, x)
\end{equation}
where $x$ is the input to the neuron.

\textit{He\_Normal} is used as the \textit{Kernel\ initializer} because we are using \textit{ReLU} activation function. The model consists of five convolution layers with \textit{filters} from 32, 64, 128, 256, and 512. \textit{Kernel\_size} of (3,3) is consistent throughout the model. \textit{Maxpooling} with a \textit{pooling\_size} of (2,2) is used in the model for dimensionality reduction and feature preservation. The final output tensor is passed through a \textit{flatten} and \textit{dense} layer before producing the final output. \textit{Batch Normalization} layers are introduced to stabilize and accelerate the training process. \textit{Adam} optimizer with a learning rate of 0.001 and \textit{clip value} of 1 is used. We use \textit{Categorical Cross Entropy} as the loss function in the model. Fig. \ref{fig1} shows the layered architecture of the CNN module.
\begin{figure}[htbp]
\includegraphics[width=1\columnwidth]{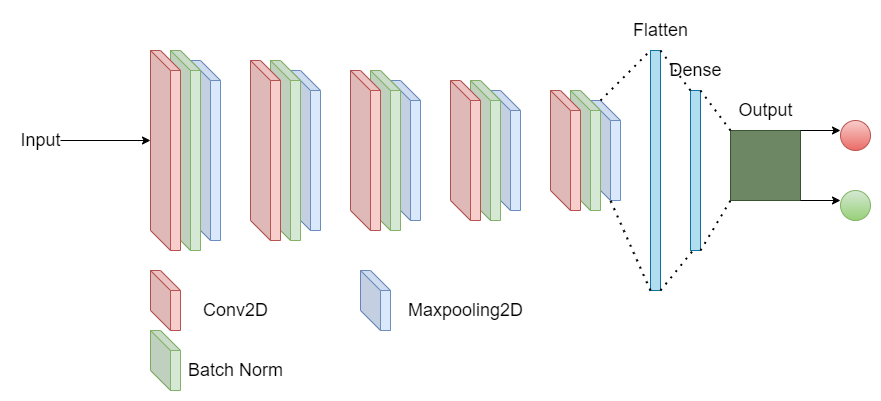}
\caption{Layered representation of the CNN module.} \label{fig1}
\end{figure}

\subsection{SE Block Attention Module}\label{sec:sebmodule}
As mentioned above, the proposed network aims to harness the capability of the SE block attention to focus on the important features of the input data by giving leverage to the channel-wise information. According to Hu et al.\cite{hu2018squeeze}, the structure of the SE block is simple and can be easily integrated with any preexisting CNN architecture with only a slight increase in computational complexity compared to the design and development of new CNN architectures. A block diagram of SE is mentioned below. The role of the SE block for extracting features varies as we go deeper into the architecture. In earlier layers, the SE block strengthens the shared low-level representations. In later layers, the SE blocks become increasingly specialized and focus on inputs in a highly class-specified manner \cite{hu2018squeeze}. The two dense layers in the SE block module has the \textit{ReLU} and \textit{Sigmoid} activation functions, respectively. \textit{He\_normal} is used as the kernel initializer. In the Squeeze phase, the \textit{Global Average Pooling (GAP)} computes the average value of each channel, resulting in a single numeric value per channel. The output is then reshaped to match the dimensions of the shared layers. The Excitation phase first applies a dense layer with \textit{ReLU} activation to the squeezed features and then it passes through the other dense layer with \textit{Sigmoid} activation in it. The final recalibrated feature is obtained by element-wise multiplication of the original input and the output of the excitation unit. An overview of the SE block operation is shown in Fig. \ref{fig2}. The mathematical descriptions for all the operations in the SE block module are taken from \cite{hu2018squeeze} are given below:

The squeeze operation is as follows, in symbols:
\begin{equation}
    z_c = \frac{1}{H \times W} \sum_{i=1}^{H} \sum_{j=1}^{W} u_c(i, j)
\end{equation}
where, $\textit{z}_c$ is the squeezed value for channel $\textit{c}$, $\textit{H}$ and $\textit{W}$ are the height and width of the input feature map, respectively, and $\textit{u}_c(i,j)$ represents the value at position $(i,j)$ in channel $\textit{c}$.

The reduce operation equation is as follows, in symbols:
\begin{equation}
    \textbf{w} = \text{ReLU}(W_1z + b_1)
\end{equation}
where, $\textbf{w}$ is the reduced dimensionality vector, $\textit{W}_1$ and $\textit{b}_1$ are the weights and biases for the reduction layer, $\textit{z}$ is the input channel descriptor from the squeeze operation, $\textit{ReLU}$ refers to the rectified linear unit activation function.

The excitation equation is as follows, in symbols:
\begin{equation}
    s = \sigma(W_2w + b_2)
\end{equation}
where, $\textit{s}$  is the channel-wise recalibration weight vector, $\textit{W}_2$ and $\textit{b}_2$ are the weights and biases for the excitation layer, $\textit{$\sigma$}$ denotes the sigmoid function.

Lastly, the scaling operation is as follows, in symbols:
\begin{equation}
    \text{\textit{v}}_c(i, j) = s_c \cdot u_c(i, j)
\end{equation}
where, $\text{\textit{v}}_c(i,j)$ is the value at position $\text{\textit{(i,j)}}$ in channel $\text{\textit{c}}$ of the output feature map, and $\text{\textit{s}}_c$ is the recalibration weight for channel $\text{\textit{c}}$.

\begin{figure}[htbp]
\includegraphics[width=1\columnwidth]{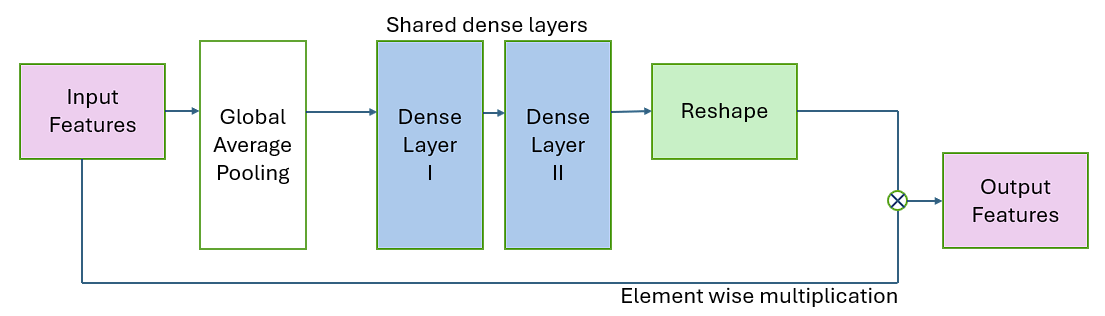}
\caption{A Squeeze-and-Excitation Block operation overview} \label{fig2}
\end{figure}
\subsection{Our Proposed Method}\label{sec:proposed-methodology}
We propose the sequential CNN with SE block attention. Our method consists of taking images and feeding them into a sequential CNN model with SE block attention integrated into the model in the deeper layers. The images pass through the convolution layers and then are passed through the SE block module to generate feature maps. In our model architecture, we have integrated the SE block with four \textit{Conv\_2D} layers. The SE block is added after the \textit{Batch\_Normalization} layer, the output of the SE block passes through the \textit{Maxpooling} layer before the feature maps are passed to the next convolution layer. This is an iterative process. Finally, the feature maps are passed through a \textit{flatten} and \textit{dense} layer before the final classification layer. The images are prepocessed and augmented before they are fed into the model for training, as shown in Fig. \ref{fig3}. The final layer in the SE block attention module uses the scaling operation to adjust the values of each channel's feature map. This configuration allows our model to focus on important and specific features across the channels more effectively. The adaptability of our model enables it to adjust the contributions of the different features dynamically based on learned importance. Our model adapts to the facial image input and learns to focus on the important features like skin textures which enables the model to effectively classify fake face images from the real ones.

\begin{figure}[htbp]
\includegraphics[width=1\columnwidth]{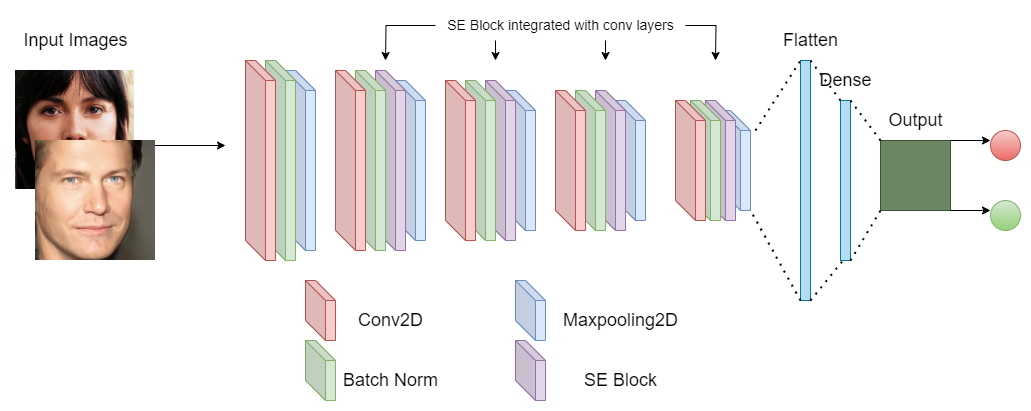}
\caption{Overview of overall model architecture.} \label{fig3}
\end{figure}
\section{Experiments and Analysis}\label{sec:experiments}
This section focuses on assessing the effectiveness and performance of our proposed techniques. Through this experimental analysis, our goal is to showcase the benefits of our methodology, particularly in terms of model evaluation. The experiments conducted, results and analysis are discussed in this section. First, the datasets and the implementation details are introduced in the next subsection. The experimental results and analysis are presented in the subsection afterward.
\subsection{Dataset and Implementation} \label{sec:datasets}
The DFFD dataset contains images generated from multiple approaches, along with actual images, which comprises approximately 300,000 images labelled as either `real' or `fake', sourced from various origins \cite{dang2020detection}. For this study, we selected fake images generated using Style GAN and real images from the Flickr-Faces-HQ dataset \cite{karras2019style} available in the DFFD dataset. The PNG format images are at 299 * 299 resolution and contain a considerable variation in terms of age, ethnicity and image background. The style-based generator focuses on pose, identity of the human faces and attributes such as freckles and hair. Samples of this dataset are shown in Fig. \ref{fig4} below.
\begin{figure}
\includegraphics[width=1\columnwidth]{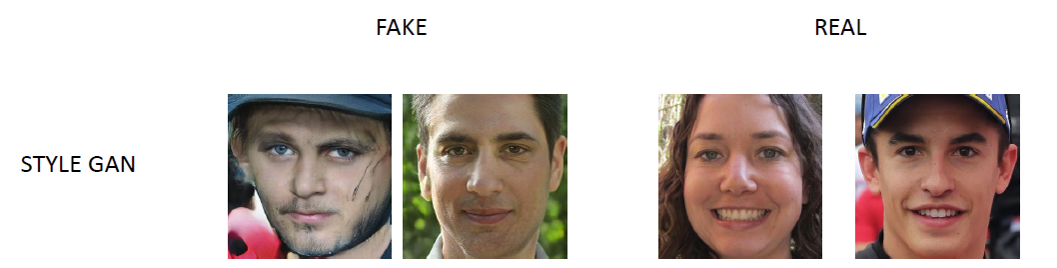}
\caption{Samples of the images from Style GAN dataset \cite{karras2019style,dang2020detection}} \label{fig4}
\end{figure}

The experiments are conducted on a 64-bit Ubuntu 22.04.3 LTS system that has an Intel Xeon(R) with a maximum memory of 512 gigabytes and 18 cores. The system includes NVIDIA Quadro RTX 8000 GPU with a memory of 48 gigabytes each.

The total number of images used for this experiment was 38,000, with an equal distribution of 19,000 images for both \textit{real} and \textit{fake} categories. The images were resized to 224 * 224 to prepare the images to feed into the network.  After resizing the images, the pixel values were re-scaled and saved as \textit{numpy} arrays. We label the \textit{fake} images as `0' and \textit{real} images as `1' and save them along with the images. Finally, we use \textit{train\_test\_split} from \textit{sk\_learn} to split the \textit{numpy} arrays into 80\% for training and 20\% for testing. \textit{Adam} optimizer is used with a lower \textit{learning rate} of 0.0001 and a \textit{Clip value} of 1.0. \textit{Categorical cross entropy} loss function is used with a \textit{batch size} of 64. The model is trained for 100 \textit{epochs} with 10-fold cross-validation.
\subsection{Results and Discussion} \label{sec:results}
 The Deepfake detection task was conducted using the sequential SE block model with facial images taken from the Style GAN dataset. In this study, we analysed the performance of the model based on its detection accuracy. Figs. 5, 6 and 7 show the detailed report of the results obtained after evaluating the model with the test dataset. These results consist of the classification report that includes the \textit{precision, recall,} and the \textit{f1-score}. The confusion matrix gives an idea about the true, false positive, and negative predictions, as well as the area under the receiver operating characteristic AUC-ROC curve.
We will show a few results of correctly and incorrectly predicted images on the test dataset.
\begin{figure}
\includegraphics[height= 4cm, width=1\columnwidth]{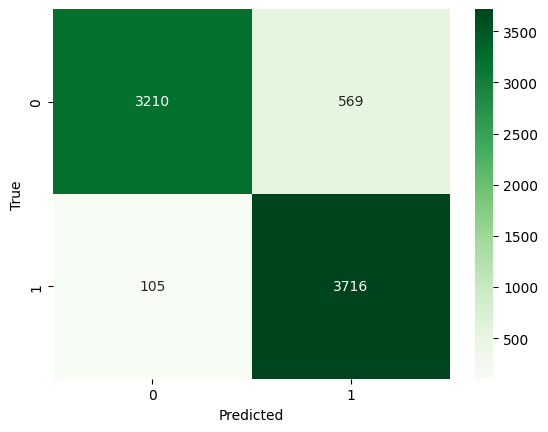}
\caption{The Confusion Matrix (Sequential)}\label{fig6}
\end{figure}
\begin{figure}
\includegraphics[height= 4cm, width=1\columnwidth]{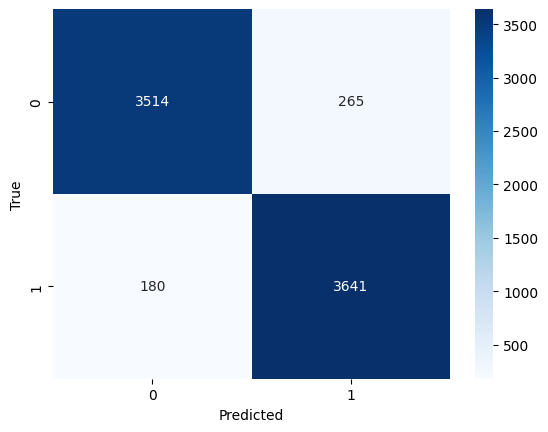}
\caption{The Confusion Matrix (Sequential + SE) }
\label{fig7}
\end{figure}

We utilized the confusion matrix to assess the performance of our model, which provides a detailed breakdown of the model`s predictions in comparison to the true labels. Fig. 5 and Fig. 6 show the confusion matrix for the Sequential CNN and our Sequential + SE model. 

 \begin{table}[htbp]
\caption{Classification report results from the experiments conducted.}
\begin{center}
\begin{tabular}{|c|c|c|c|c|c|c|c|}
\hline
\textbf{Model}&\multicolumn{2}{|c|}{\textbf{Precision}} &\multicolumn{2}{|c|}{\textbf{Recall}}&\multicolumn{2}{|c|}{\textbf{F1-score}}& \textbf{Accuracy} \\
\hline
\textbf{} & \textbf{\textit{F}}& \textbf{\textit{R}}&
            \textbf{\textit{F}}& \textbf{\textit{R}}& \textbf{\textit{F}}& \textbf{\textit{R}}& \\
\hline
CNN    &.97&.87&.85&.97&.91&.92&91.13\%\\
\hline
CNN+SE &.95&.93&.93&.95&.94&.94&94.14\%\\

\hline
\multicolumn{4}{l}{\textit{F:Fake, R:Real}}
\end{tabular}
\label{tab1}
\end{center}
\end{table}

The other metrics for model evaluation such as \textit{precision}, \textit{recall} and \textit{F1-score} are provided in Table I. Our model demonstrates commendable performance in distinguishing between `Fake' and `Real' categories, achieving precision scores of 0.95 and 0.93, respectively. Notably, the model balances recall effectively, with scores of 0.93 for `Fake' and 0.95 for `Real', leading to high F1-scores of 0.94 and 0.94, indicating a strong harmony between precision and recall across both categories. We have conducted the experiments on the sequential model without the SE block module. The results are also provided in Table I.
\begin{figure}
\centering  
\includegraphics[height= 4cm, width=1\columnwidth]{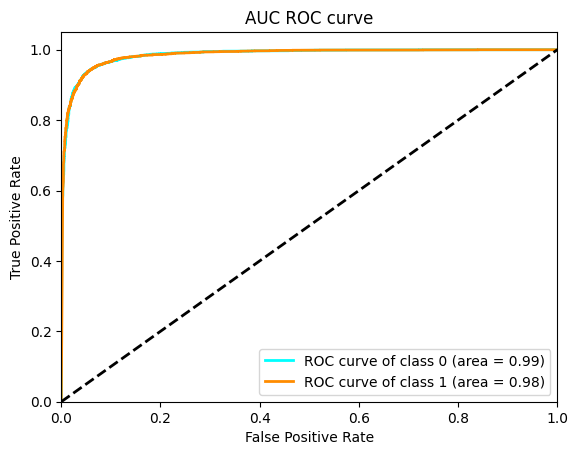}
\caption{AUC-ROC curve of the (Sequential + SE) model. AUC-ROC score of 0.99 for Class 0 (`fake'), and AUC-ROC score of 0.98 for class 1 (`real') }
\label{fig7}
\end{figure}

In the evaluation of our binary classification model, aimed at distinguishing between `fake' (Class 0) and `real' (Class 1) instances, we achieved outstanding performance as evidenced by the (ROC) curve (AUC-ROC) scores Fig.\ref{fig7}. Specifically, the model demonstrated a near-perfect ability to identify `fake' instances with an AUC-ROC score of 0.99 for Class 0 and an AUC-ROC score of 0.98 for class 1 `real' instances. This suggests that our model has 98\% probability of accurately classifying `real' versus `fake' instances under the same conditions. This level of accuracy is indicative of our model's sophisticated feature processing and classification mechanisms, making it a highly reliable tool in scenarios where distinguishing between `Fake' and `Real' is critical.

Similar work of detecting Deepfake using CNN and self-attention module was done by Huebens et al. where the model reported an AUC-ROC score of 90.9 on the DFDC dataset \cite{hubens2021fake}. Our model uses the SE block attention module with CNN and achieved an AUC-ROC score of 98.5 on the Style GAN dataset from DFFD. A comparison of our method with some existing methods on different datasets are shown in Table II.
\begin{table}[htbp]
\caption{Comparison of the performance of the model with some existing detection methods}
\begin{center}
\begin{tabular}{|c|c|c|c|c|}
\hline
\textbf{Model}&{\textbf{Dataset}}&{\textbf{AUC}}&\textbf{Model Size(MB)} \\
\hline
Fake-buster(SA) \cite{hubens2021fake} &DFDC&90.9&49.17\\
\hline
CNN \cite{pryor2023deepfake} &140k&88.33&-\\
\hline
SVM \cite{pryor2023deepfake} &140k&81.69&-\\
\hline
WM \cite{cuihaoleo_kaggle-dfdc} &DFDC&98.5&1722.85\\
\hline
\textbf{\textit{Proposed(CNN+SE)}}&Style GAN(DFFD)&\textbf{98.5}&56.18\\
\hline
\end{tabular}
\label{tab1}
\end{center}
\end{table}

As seen in Table II, our CNN with SE block demonstrates superior performance compared to existing methods such as Fake-buster (SA) \cite{hubens2021fake} and prior detection methods using CNN and SVM \cite{pryor2023deepfake}, as evidenced by higher AUC scores. Moreover, our model achieves comparable AUC-ROC scores to the model presented by WM\cite{cuihaoleo_kaggle-dfdc}, while significantly reducing model size from 1722.85 MB to 56.18 MB, indicating efficient parameter utilization without sacrificing performance.

In the evaluation of our proposed model's performance on the test dataset, we provide a selection of sample images illustrating both correct and incorrect predictions in Figs. 8 and 9. These examples offer valuable insights into the model's capabilities in distinguishing between `fake' and `real' instances, while also shedding light on scenarios where the model may encounter challenges.

Our proposed model accurately classifies `fake' and `real' images, identifying subtle inconsistencies in texture, lighting, and facial features. This highlights the effectiveness of the SE block attention mechanism in capturing spatial and channel-wise information. However, incorrect predictions indicate potential limitations, such as incorrect predictions of fake images as real or vice versa, indicating challenges in capturing nuanced visual cues or handling complex manipulations. These incorrect predictions provide valuable insights into specific scenarios or image characteristics, guiding future refinements or optimizations to enhance performance.
\begin{figure}[htbp]
\includegraphics[height=2cm, width=\columnwidth]{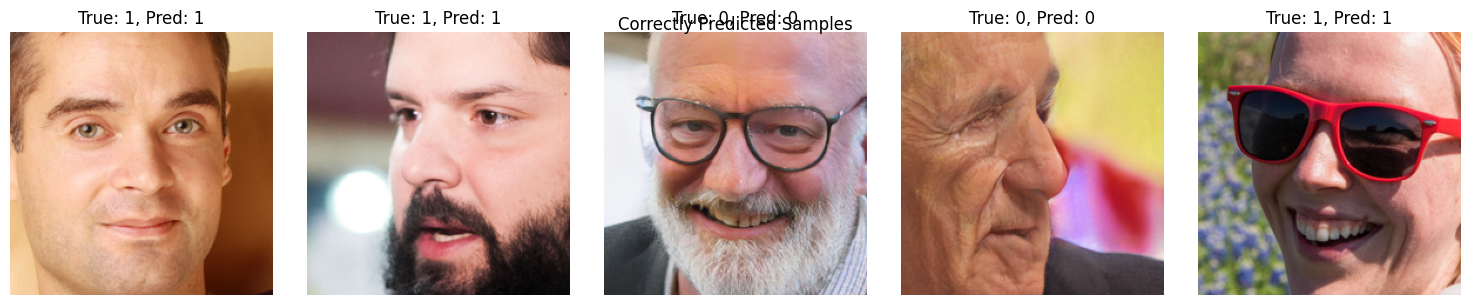}
\caption{Correct Predictions of our model on test data} \label{fig8}
\end{figure}
\begin{figure}[h]
\includegraphics[height=2cm, width=\columnwidth]{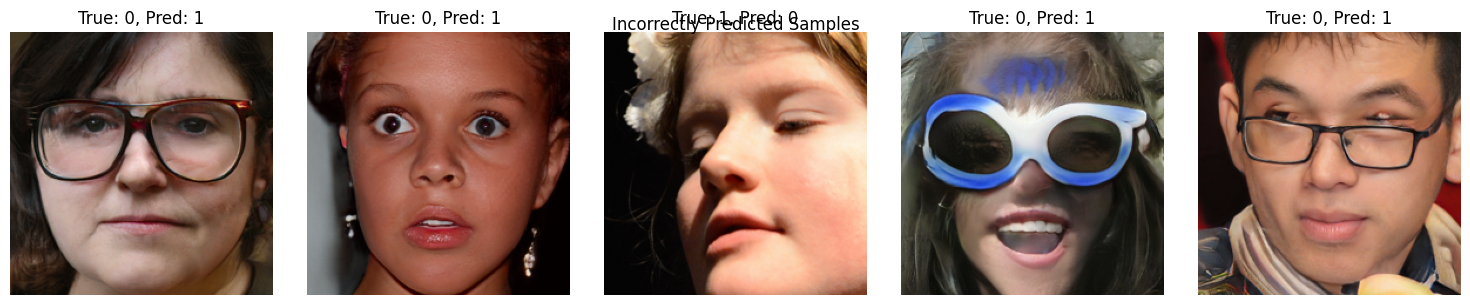}
\caption{Incorrect Predictions of our model on test data} \label{fig9}
\end{figure}

\section{Conclusion}\label{sec:conclusion}
We developed a model to detect Deepfake images using a simple CNN model with SE block attention. Our model produced a good level of accuracy on correctly predicting the \textit{real} and \textit{fake} images. Labelled data was utilized to conduct image classification. We evaluated our model along with a Sequential CNN model without SE block attention. SE block model demonstrated superior accuracy. The enhanced model proves effective in identifying manipulated images, highlighting its improved performance and detection capabilities. Our model showed better performance in classification accuracy and AUC-ROC scores when compared to the previous research work incorporating the Self Attention module, and different deep learning architectures.

In the future, we plan to assess various datasets and other attention modules like Convolution Block Attention Module (CBAM) and scaled dot product attention. Evaluating different models will provide a better indication of the utilization of attention mechanisms with CNN. Inclusion of more datasets will give a better estimates of the model performance. Different combinations of hyperparameters and hyperparameter tuning can be performed to improve the prediction accuracy of the model. Also, we further plan to incorporate explainability to the models as well. The outputs of the models can be compared and analyzed to study Deepfake so that the detection of Deepfake can be done in the most efficient way using a lightweight framework.

\section*{Acknowledgment}
This research is funded by the National Science Foundation (NSF), Award number: 1900187.
\bibliographystyle{IEEEtran}  
\input{conference_101719.bbl}


\end{document}

%% file: conference_101719.bbl

%% file: conference_101719.bbl
\begin{thebibliography}{10}
\providecommand{\url}[1]{#1}
\csname url@samestyle\endcsname
\providecommand{\newblock}{\relax}
\providecommand{\bibinfo}[2]{#2}
\providecommand{\BIBentrySTDinterwordspacing}{\spaceskip=0pt\relax}
\providecommand{\BIBentryALTinterwordstretchfactor}{4}
\providecommand{\BIBentryALTinterwordspacing}{\spaceskip=\fontdimen2\font plus
\BIBentryALTinterwordstretchfactor\fontdimen3\font minus \fontdimen4\font\relax}
\providecommand{\BIBforeignlanguage}[2]{{%
\expandafter\ifx\csname l@#1\endcsname\relax
\typeout{** WARNING: IEEEtran.bst: No hyphenation pattern has been}%
\typeout{** loaded for the language `#1'. Using the pattern for}%
\typeout{** the default language instead.}%
\else
\language=\csname l@#1\endcsname
\fi
#2}}
\providecommand{\BIBdecl}{\relax}
\BIBdecl

\bibitem{mirsky2021creation}
Y.~Mirsky and W.~Lee, ``The creation and detection of deepfakes: A survey,'' \emph{ACM Computing Surveys (CSUR)}, vol.~54, no.~1, pp. 1--41, 2021.

\bibitem{yu2021survey}
P.~Yu, Z.~Xia, J.~Fei, and Y.~Lu, ``A survey on deepfake video detection,'' \emph{Iet Biometrics}, vol.~10, no.~6, pp. 607--624, 2021.

\bibitem{seow2022comprehensive}
J.~W. Seow, M.~K. Lim, R.~C. Phan, and J.~K. Liu, ``A comprehensive overview of deepfake: Generation, detection, datasets, and opportunities,'' \emph{Neurocomputing}, vol. 513, pp. 351--371, 2022.

\bibitem{kong2022digital}
C.~Kong, S.~Wang, H.~Li \emph{et~al.}, ``Digital and physical face attacks: Reviewing and one step further,'' \emph{APSIPA Transactions on Signal and Information Processing}, vol.~12, no.~1, 2022.

\bibitem{huang2023implicit}
B.~Huang, Z.~Wang, J.~Yang, J.~Ai, Q.~Zou, Q.~Wang, and D.~Ye, ``Implicit identity driven deepfake face swapping detection,'' in \emph{Proceedings of the IEEE/CVF Conference on Computer Vision and Pattern Recognition}, 2023, pp. 4490--4499.

\bibitem{naitali2023deepfake}
A.~Naitali, M.~Ridouani, F.~Salahdine, and N.~Kaabouch, ``Deepfake attacks: Generation, detection, datasets, challenges, and research directions,'' \emph{Computers}, vol.~12, no.~10, p. 216, 2023.

\bibitem{rossler2019faceforensics++}
A.~Rossler, D.~Cozzolino, L.~Verdoliva, C.~Riess, J.~Thies, and M.~Nie{\ss}ner, ``Faceforensics++: Learning to detect manipulated facial images,'' in \emph{Proceedings of the IEEE/CVF international conference on computer vision}, 2019, pp. 1--11.

\bibitem{dang2020detection}
H.~Dang, F.~Liu, J.~Stehouwer, X.~Liu, and A.~K. Jain, ``On the detection of digital face manipulation,'' in \emph{Proceedings of the IEEE/CVF Conference on Computer Vision and Pattern recognition}, 2020, pp. 5781--5790.

\bibitem{zhu2021attention}
B.~Zhu, P.~Hofstee, J.~Lee, and Z.~Al-Ars, ``An attention module for convolutional neural networks,'' in \emph{Artificial Neural Networks and Machine Learning--ICANN 2021: 30th International Conference on Artificial Neural Networks, Bratislava, Slovakia, September 14--17, 2021, Proceedings, Part I 30}.\hskip 1em plus 0.5em minus 0.4em\relax Springer, 2021, pp. 167--178.

\bibitem{wodajo2021deepfake}
D.~Wodajo and S.~Atnafu, ``Deepfake video detection using convolutional vision transformer,'' \emph{arXiv preprint arXiv:2102.11126}, 2021.

\bibitem{hu2018squeeze}
J.~Hu, L.~Shen, and G.~Sun, ``Squeeze-and-excitation networks,'' in \emph{Proceedings of the IEEE conference on computer vision and pattern recognition}, 2018, pp. 7132--7141.

\bibitem{rana2022deepfake}
M.~S. Rana, M.~N. Nobi, B.~Murali, and A.~H. Sung, ``Deepfake detection: A systematic literature review,'' \emph{IEEE access}, vol.~10, pp. 25\,494--25\,513, 2022.

\bibitem{dosovitskiy2020image}
A.~Dosovitskiy, L.~Beyer, A.~Kolesnikov, D.~Weissenborn, X.~Zhai, T.~Unterthiner, M.~Dehghani, M.~Minderer, G.~Heigold, S.~Gelly \emph{et~al.}, ``An image is worth 16x16 words: Transformers for image recognition at scale,'' \emph{arXiv preprint arXiv:2010.11929}, 2020.

\bibitem{coccomini2022combining}
D.~A. Coccomini, N.~Messina, C.~Gennaro, and F.~Falchi, ``Combining efficientnet and vision transformers for video deepfake detection,'' in \emph{International conference on image analysis and processing}.\hskip 1em plus 0.5em minus 0.4em\relax Springer, 2022, pp. 219--229.

\bibitem{wu2021cvt}
H.~Wu, B.~Xiao, N.~Codella, M.~Liu, X.~Dai, L.~Yuan, and L.~Zhang, ``Cvt: Introducing convolutions to vision transformers,'' in \emph{Proceedings of the IEEE/CVF international conference on computer vision}, 2021, pp. 22--31.

\bibitem{simonyan2014very}
K.~Simonyan and A.~Zisserman, ``Very deep convolutional networks for large-scale image recognition,'' \emph{arXiv preprint arXiv:1409.1556}, 2014.

\bibitem{roy20223d}
R.~Roy, I.~Joshi, A.~Das, and A.~Dantcheva, ``3d cnn architectures and attention mechanisms for deepfake detection,'' in \emph{Handbook of Digital Face Manipulation and Detection: From DeepFakes to Morphing Attacks}.\hskip 1em plus 0.5em minus 0.4em\relax Springer International Publishing Cham, 2022, pp. 213--234.

\bibitem{karras2019style}
T.~Karras, S.~Laine, and T.~Aila, ``A style-based generator architecture for generative adversarial networks,'' in \emph{Proceedings of the IEEE/CVF conference on computer vision and pattern recognition}, 2019, pp. 4401--4410.

\bibitem{hubens2021fake}
N.~Hubens, M.~Mancas, B.~Gosselin, M.~Preda, and T.~Zaharia, ``Fake-buster: A lightweight solution for deepfake detection,'' in \emph{Applications of Digital Image Processing XLIV}, vol. 11842.\hskip 1em plus 0.5em minus 0.4em\relax SPIE, 2021, pp. 146--154.

\bibitem{pryor2023deepfake}
L.~Pryor, R.~Dave, M.~Vanamala \emph{et~al.}, ``Deepfake detection analyzing hybrid dataset utilizing cnn and svm,'' \emph{arXiv preprint arXiv:2302.10280}, 2023.

\bibitem{cuihaoleo_kaggle-dfdc}
H.~Cui, ``{kaggle-dfdc: Deepfake Detection Challenge (DFDC) solution},'' \url{https://github.com/cuihaoleo/kaggle-dfdc}, 2022, accessed: April 2024.

\end{thebibliography}
